\documentclass{article}

\usepackage[preprint]{neurips_2020}

\usepackage[utf8]{inputenc} 
\usepackage[T1]{fontenc}    
\usepackage{hyperref}       
\usepackage{url}            
\usepackage{booktabs}       
\usepackage{amsfonts}       
\usepackage{nicefrac}       
\usepackage{microtype}      
\usepackage{graphicx}
\usepackage{algorithmic}
\usepackage[linesnumbered,ruled, norelsize]{algorithm2e}
\usepackage{booktabs}
\usepackage{algorithmic}
\usepackage[fleqn]{amsmath}

\title{Additively Homomorphical Encryption based Deep Neural Network for Asymmetrically Collaborative Machine Learning}

\author{%
  Yifei Zhang \\
  Chinese University of Hong Kong\\
  \texttt{yifeiacc@gmail.com} \\
  \And
  Hao Zhu \\
  Australian National University \\
  \texttt{allenhaozhu@gmail.com} \\
}

\begin{document}

\maketitle

\begin{abstract}
The financial sector presents many opportunities to apply various machine learning techniques. Centralized machine learning creates a constraint which limits further applications in finance sectors. Data privacy is a fundamental challenge for a variety of finance and insurance applications that account on learning a model across different sections.  In this paper, we define a new practical scheme of collaborative machine learning that one party owns data, but another party owns labels only, and term this \textbf{Asymmetrically Collaborative Machine Learning}. For this scheme, we propose a novel privacy-preserving architecture where two parties can collaboratively train a deep learning model efficiently while preserving the privacy of each party's data.  More specifically, we decompose the forward propagation and backpropagation of the neural network into four different steps and propose a novel protocol to handle information leakage in these steps. Our extensive experiments on different datasets demonstrate not only stable training without accuracy loss, but also more than 100 times speedup compared with the state-of-the-art system. 
\end{abstract}

\section{Introduction}
Machine learning makes use of algorithms to perform tasks such as prediction or classification. 
Classical machine learning approaches need centralizing the training data on one machine or in a cluster.
Centralized machine learning is, by far, the most common architecture.
However, it also stops further applications in finance sectors (e.g., insurance and bank) from employing machine learning techniques.
For confidentiality reasons, these finance sectors are not able to share their data and store it in the cloud, and thus cannot benefit from centralized machine learning with other organizations.

Data privacy is a fundamental challenge for many machine learning applications depending on data aggregation across different entities, especially in finance sectors. 
The trade-off between data privacy and learning on aggregated data creates a collaborative circumstance. 
Decentralized machine learning applies to keeps data safe and ensures privacy under some conditions. 
A new method called Federated Learning (or Collaborative Machine Learning) \cite{mcmahan2017communication}\cite{yang2019federated}, proposed by Google, offers a way of decentralized and confidential machine learning.
By using aggregated updating parameters of the model to train algorithms instead of raw data, federated learning empowers sectors where data cannot be transferred to third parties for confidentiality reasons with data network effects.
In fact, federated learning plays a similar role in data parallelism based distributed machine learning. In other words, federated learning is a kind of distributed version of deep neural networks that the data can be partitioned and stored in multiple machines.  
However, federated learning is only able to transform (or online) deep learning into horizontal parallelism, which splits the data based on the quantity of the data, i.e., the different amount of data subset goes into the parallel computation. Thus, the learning scheme of federated learning cannot handle the situation of vertical parallelism. Additionally, recent researches~\cite{zhu2019deep} have shown that it is possible to obtain the private training data from the publicly shared gradients.


In this paper, we investigate a learning technique that allows two parties to collaboratively train a model while one party (the active party) holds data and another party (the passive party) holds the corresponding labels. We term this \textbf{Asymmetrically Collaborative Machine Learning} since the learning task is based on the model that parties actively interact by sharing information and take on asymmetric roles. Compared with federated learning and collaborative machine learning, asymmetrically collaborative machine learning is to transform machine learning algorithms into Vertical
parallelism, which splits the data based on one or more specific internal characteristics of the data.

Technically, for this scheme, the straightforward solution is learning on encrypted data. There are two different ways to learn a model on encrypted data: differential privacy~\cite{dwork2014algorithmic,abadi2016deep} and homomorphic encryption~\cite{acar2018survey}. Differential privacy injects noise into query results to avoid inferring information about any specific record. However, it needs careful calibration to balance privacy and model usability. Further, private attributes still remain in plaintext, which are unacceptable for finance sectors, so users may still have security concerns. %
A more promising solution comes from the recent advance in homomorphic encryption. It allows users to encrypt data with the public key and offload computation to the cloud (or other parties). The cloud computes on the encrypted data and generates encrypted results. Without the secret key,  the cloud simply serves as a computation platform but cannot access any user information. But it is extremely costly in computation, thus unsuitable for high dimension data and costly machine learning methods (e.g., deep neural networks).

In this paper, we propose a novel privacy-preserving architecture to solve the problem of how the two parties can collaboratively train a model while one party holds data, and another party holds the label by neural networks. 
Intuitively, we avoid learning on encrypted data directly. Thus we decompose a Deep Neural Network (DNN) into two components: the 'Feature Extraction' part and the 'Classifier' part. In 'Feature Extraction' a locally unencrypted deep neural network is used to extract more compact features from unencrypted data. In the 'Classifier,' a shallow neural network is used to learn a classifier on encrypted features came from the component of feature extraction.
However, such a locally unencrypted and locally encrypted deep neural network results in a new challenge of preserving-privacy. 
Compared with learning on encrypted data, the 'classifier' needs to calculate the gradient of inputs and passes it back to the 'feature extraction' part. Thus we also need to avoid information leakage in the gradient, which has been proved in \cite{zhu2019deep}. 
Furthermore, we take a two-layer neural network as an example to decompose the forward and backward propagation into four different steps and propose a protocol to avoid information leakage in them.
%
Our contribution includes:
\begin{itemize}
    \item We define a new scheme of collaborative machine learning, called asymmetrically collaborative machine learning which is promising in many real-world applications.
    
    \item We propose a novel privacy-preserving, computationally efficient, homomorphic encryption-based backpropagation algorithm for asymmetrically collaborative machine learning.
    
    \item An extensive empirical evaluation of the proposed approach is demonstrated to show the proposed method achieves 100 times speed-up compared with the methods based on learning on encrypted data.
\end{itemize}


\section{Related Work}

Traditional encryption methods, such as AES (Advanced Encryption Standard) \cite{daemen1999aes}, are extremely fast, and allow data to be stored conveniently in encrypted form. However, it is costly and quite a challenge to perform even simple analytics on the encrypted data (i.e. ciphertexts). For example, the cloud server needs access to the secret key, and the owner of the data needs to download, decrypt as well as operate on the data locally. These commonly lead to security concerns. Homomorphic Encryption (HE) is able to encrypt data before sending to a cloud computing platform while still allowing the operations of search, sort, and edit on the ciphertexts. This property avoids needing to ship data back and forth to be decrypted from the cloud computing platform. HE which supports any function on ciphertexts is known as Fully Homomorphic Encryption (FHE)~\cite{gentry2009fully}, while Partially Homomorphic Encryption (PHE)~\cite{damgaard2012multiparty,juvekar2018gazelle} includes encryption schemes that have homomorphic properties, with respect to one operation (e.g., only addition or only multiplication, but not both). However, FHE faces a fundamental problem which extremely costly in computation. Thus it is unpractical for some machine learning methods (e.g. high dimension linear model and neural network).
Paillier~\cite{paillier1999public}, as well known one method of PHE, supports unlimited numbers of additions between ciphertext, and multiplication between a ciphertext and a scalar constant. In other words, given $Enc(m_1)$ and $Enc(m_2)$, you can not get $Enc(m_1)\otimes Enc(m_2)$. You can only get $Enc(m_1)\oplus Enc(m_2)$ which equals $Enc(m_1+ m_2)$. Given $Enc(m_1)$ and $m_2$, you can get $Enc(m_1)\otimes m_2$ which equals $Enc(m_1\cdot m_2)$. But notice that $m_2$, in this case, was not encrypted. $\oplus$ is the homomorphic addition with ciphertext and $\otimes$ is the homomorphic multiplication between ciphertext and a scalar constant. $Enc(x)$ is the ciphertext of plaintext $x$.

GELU-Net \cite{ijcai2018-547} proposes a novel privacy-preserving architecture based on paillier. The main difference between the proposed method and GELU is that GELU works under the situation that one party
%
In this paper, we focus on solving the privacy issue that data and labels are distributed on different parties.

\section{Methodology}

\subsection{Applying Neural Networks to Collaborative Machine Learning}
\label{sec1}
To solve this problem, our main strategy is making a deep neural network partially learn on unencrypted data and partially learn on encrypted features to reduce the computation cost in the stage of learning on encrypted data. 
More specifically, a deep neural network is carefully partitioned into two parts: the 'feature extraction' and the 'classifier' (or 'regression'). The 'feature extraction' plays a role of dimension reduction in the plaintext and produces compact features to the 'classifier'. And then the implementation of 'the classifier' is learning a simplified model (e.g. logistic regression) on encrypted features generated from the 'feature extraction' part. It is easy to observe that this setting is insensitive to the dimension of inputs and network architectures not only because the 'feature extraction' part contains the majority of neural network and it does not use arithmetic operations of homomorphic encryption, but also because the computational complexity of 'classifier' part is related to the dimension of features generated from the 'feature extraction' part rather than the dimension of raw inputs and the former is far more compact. Take a two-layer neural network as an example, one for 'feature extraction' and another one for 'classifier',
The corresponding training objective is shown in Eq.~\ref{equ:erf}, for simplicity the bias items are omitted:
\begin{align}
    \mathbf{\min_{\mathbf{W_1, W_2}}}\frac{1}{|D|}\sum_{ \mathcal{D}} \mathcal{L}_i(\mathbf{t_i}, \mathcal{F}_{\{\mathbf{W_1, W_2}\}}(\mathbf{x_i}))
        =\sum_{\mathcal{D}} \mathcal{L}_i(\mathbf{t_i}, \text{Softmax}(\mathbf{{W_2}}\sigma(\mathbf{W_1}\mathbf{x_i}))), \mathbf{x_i} \in \mathcal{D}
\label{equ:erf}
\end{align}
where $\mathcal{D}$ is the dataset. $\mathbf{x_i}$ and $\mathbf{t_i}$ represend the input data sample and its related target (label). $\sigma(.)$ is ReLU function~\cite{glorot2011deep}
$\mathbf{W_1}\in R^{d_h\times d_i}$ and $\mathbf{W_2}\in R^{c \times d_h}$ are parameters of two layers respectively. $d_i$ is the dimension of input sample. $d_h$ is the dimension of hidden layer. $c$ is the output dimension as the same as the number of categories. Assuming $\mathcal{L}_i$ is  CrossEncropy~\cite{lecun2015deep}, $\eta$ is learning rate, We can further decompose forward and backward propagation into two different parts respectively, and then there are four steps as following:
\begin{itemize}
    \item \textbf{Step 1} The forward propagation on the active party: the active party feeds the data into the neural net (feature extraction ) on the active party. The output activations 
    $
        \mathbf{a_i}=\sigma(\mathbf{W_1 x_i})\in R^d_h
    $
    is then send to the passive party.
    \item \textbf{Step 2} The forward propagation on the passive party: the passive party propagates the received activation $
    \mathbf{a_i}$ through its neural nets 
    $
        \mathbf{p_i}=\text{softmax}\mathbf{(z_i)},\mathbf{z_i}=\mathbf{W_2a_i}
    $
    and compute the CrossEncropy loss $\mathcal{L}_i(\mathbf{t_i}, \mathbf{p_i}) =  \mathbf{t_i}\log(\mathbf{p_i})$.
    
    \item \textbf{Step 3} The back propagation on the passive party: the passive party compute the required gradients:
    $\frac{\partial \mathcal{L}_i}{\partial \mathbf{p_i}} = \mathbf{t_i} - \mathbf{p_i}$, 
        $\frac{\partial \mathcal{L}_i}{\partial \mathbf{W_2}} = \frac{\partial \mathcal{L}_i}{\partial \mathbf{p_i}} \frac{\partial \mathbf{p_i}}{\partial \mathbf{W_2}} = \frac{\partial \mathcal{L}_i}{\partial \mathbf{p_i}} \mathbf{a_i}$, 
        $\frac{\partial \mathcal{L}_i}{\partial \mathbf{a_i}} = \frac{\partial \mathcal{L}_i}{\partial \mathbf{p_i}} \frac{\partial \mathbf{p_i}}{\partial \mathbf{a_i}} = \frac{\partial \mathcal{L}_i}{\partial \mathbf{p_i}} \mathbf{W_2}$
    and update parameter via $\mathbf{W_2} = \mathbf{W_2} - \eta \frac{\partial \mathcal{L}_i}{\partial \mathbf{W_2}}$ and send $\frac{\partial \mathcal{L}_i}{\partial \mathbf{a_i}}$ back to the active party.
    \item \textbf{Step 4} The back propagation on the active party: the active party receives the gradient $\frac{\partial \mathcal{L}_i}{\partial \mathbf{a_i}}$ from passives party, computes the gradient $\frac{\partial \mathcal{L}_i}{\partial \mathbf{W_1}} = \frac{\partial \mathcal{L}_i}{\mathbf{a_i}} \frac{\partial 
    \mathbf{a_i}}{\partial \mathbf{W_1}} = \frac{\partial \mathcal{L}_i}{\partial \mathbf{a_i}} \mathbf{x_i}$ 
    and updates the parameters $\mathbf{W_1} = \mathbf{W_1} - \eta \frac{\partial \mathcal{L}_i}{\partial \mathbf{W_1}}$.
\end{itemize}
To fulfil the requirement that both data and label cannot get or infer from the other side, we encrypt the intermediate results exchanged in both parties, and carefully design a Privacy-Preserving Back-Propagation that is compatible with the PHE in the setting of Asymmetrically Collaborative Machine Learning. We detail the privacy issue and the algorithm in the following section.

\subsection{Secure Forward Propagation on the active party}
The secure forward propagation on the active party is similar to the normal forward propagation used in training neural nets. The only difference is that the last layer activations (outputs) $\mathbf{a_i}$ in the secure forward propagation needs to be encrypted with PHE. We note the additive homomorphic encryption as $\mathbf{[ . ]_c}$ Since these activations will be transmitted to the passive party, leaving them in the plaintext will lead to the issue of information leakage. The passive party or attacker can collect these activations as meaningfully features for later use. Potentially, the passive party or attacker may infer the personal information from these activations. Thus, the active party will encrypt the activations to $[\mathbf{a_i}]_c$ and then send to the passive party.
\begin{algorithm}
\caption{Privacy-preserved Forward Propagation}
\label{alg:A}
\begin{algorithmic}
\STATE \textbf{Initialization:} $\epsilon_{acc}$, $\mathbf{\widetilde{W}_2}$, $\mathbf{W_1}$
\STATE {\textbf{Input:} learning rate $\eta$, data sample $\mathbf{x}$}
\STATE {\textbf{Active Party:}}
\STATE {$\mathbf{a}_i \leftarrow \sigma(\mathbf{W_2}\mathbf{x}$)}, {$[\mathbf{a}_i]_c \leftarrow Enc(\mathbf{a}_i)$}
\STATE Send $[\mathbf{a}_i]_c$ to passive party
\STATE {\textbf{Passive Party:}}
\STATE Compute the weighted sum $[\widetilde{\mathbf{z_i}}]_c \leftarrow \mathbf{\widetilde{W}_2}\bigotimes\mathbf{[a_i]_c}$ and add random noise $[\widetilde{\mathbf{z_i}} + \epsilon_s]_c \leftarrow \epsilon_s \oplus [\mathbf{z_i}]_c$
\STATE Sent $[\widetilde{\mathbf{z_i}} + \epsilon_s]_c$ to active party
\STATE {\textbf{Active Party:}}
\STATE {$\widetilde{\mathbf{z_i}} + \epsilon_s \leftarrow$ $Dec([\widetilde{\mathbf{z_i}} + \epsilon_s]_c)$ }
\STATE Remove noise: $\mathbf{z_i} + \epsilon_s \leftarrow \widetilde{\mathbf{z_i}} - \epsilon_{acc}\mathbf{a}_i + \epsilon_s$\;
\STATE Send $\mathbf{z_i} + \epsilon_s$ to the passive party
\STATE {\textbf{Passive Party:}}
\STATE Remove noise:  $\mathbf{z_i}  \leftarrow \mathbf{z_i} + \epsilon_s$
\STATE Compute softmax result $\mathbf{p_i} = Softmax(\mathbf{z_i})$ 
\STATE \textbf{Return} $\mathbf{p_i}$ 
\end{algorithmic}
\end{algorithm}
\begin{algorithm}[h]
\caption{Privacy-preserved Backward Propagation}
\SetAlgoLined
\SetKwInOut{Input}{Input}
\SetKwInOut{Output}{Output}
\SetKw{AP}{Active Party}
\SetKw{PP}{Passive Party}
\begin{algorithmic}

\STATE 
\textbf{Input:} Prediction: $\mathbf{p_i}$ on passive party; Target $\mathbf{t_i}$; The encrypted activation $[\mathbf{a_i}]_c$\;
\PP{\;}
Compute the following gradients:\;
$\frac{\partial \mathcal{L}_i}{\partial \mathbf{p_i}} \leftarrow \mathbf{t_i} - \mathbf{p_i}, [\frac{\partial \mathcal{L}_i}{\partial \mathbf{W_2}}]_c  \leftarrow \frac{\partial \mathcal{L}_i}{\partial \mathbf{p_i}}\bigotimes [\mathbf{a_i}]_c, \widetilde{\frac{\partial \mathcal{L}_i}{\partial \mathbf{a_i}}}  \leftarrow \frac{\partial \mathcal{L}_i}{\partial \mathbf{p_i}} {\mathbf{\widetilde{W}_2}}$\;
Add noise $[\frac{\partial \mathcal{L}_i}{\partial \mathbf{W_2}} + \epsilon_s]_c \longleftarrow  [\frac{\partial \mathcal{L}_i}{\partial \mathbf{W_2}}]_c \bigoplus \epsilon_s $ \;
Send $[\frac{\partial \mathcal{L}_i}{\partial \mathbf{W_2}} + \epsilon_s]_c$ to the active party \;
\AP{\;}
$\frac{\partial \mathcal{L}_i}{\partial \mathbf{W_2}} + \epsilon_s \longleftarrow Dec([\frac{\partial \mathcal{L}_i}{\partial \mathbf{W_2}} + \epsilon]_c)$\;
Add noise: $ \widetilde{\frac{\partial \mathcal{L}_i}{\partial \mathbf{W_2}}}+ \epsilon_s \longleftarrow {\frac{\partial \mathcal{L}_i}{\partial \mathbf{W_2}}} + \epsilon_s -\frac{{\epsilon_w}}{\eta}$\;
Encrypt noise: $[\epsilon_{acc}]_c\longleftarrow Enc(\epsilon_{acc})$\;
Accumulate noise: $\epsilon_{acc} \leftarrow \epsilon_{acc} + \epsilon_w$\;
Send $ \widetilde{\frac{\partial \mathcal{L}_i}{\partial \mathbf{W_2}}}+ \epsilon_s$ and $[\epsilon_{acc}]_c$ to passive party.\;
\PP{\;}
Remove noise: $ \widetilde{\frac{\partial \mathcal{L}_i}{\partial \mathbf{W_2}}} \leftarrow  \widetilde{\frac{\partial \mathcal{L}_i}{\partial \mathbf{W_2}}}+ \epsilon_s$\;
Update weights ${\mathbf{\widetilde{W}_2}} \leftarrow {\mathbf{\widetilde{W}_2}} - \eta  \widetilde{\frac{\partial \mathcal{L}_i}{\partial \mathbf{W_2}}}$\;
Remove noise from gradients: $[\frac{\partial \mathcal{L}_i}{\partial \mathbf{a_i}}]_c \leftarrow \widetilde{\frac{\partial \mathcal{L}_i}{\partial \mathbf{a_i}}} - [\epsilon_{acc}]_c \bigotimes \frac{\partial \mathcal{L}_i}{\partial \mathbf{p_i}}$\;
Send $[\frac{\partial \mathcal{L}_i}{\partial \mathbf{a_i}}]_c$ to the active party\;
\AP{\\}
$\frac{\partial \mathcal{L}_i}{\partial \mathbf{a_i}} \longleftarrow Dec([\frac{\partial \mathcal{L}_i}{\partial \mathbf{a_i}}]_c)$\;
Do the normal backpropagation with $\frac{\partial \mathcal{L}_i}{\partial \mathbf{a_i}}$\;
\end{algorithmic}
\end{algorithm}

\subsection{Secure Forward Propagation on the passive party}
\label{sec3}
After receiving the encrypted activation $[\mathbf{a_i}]_c$, the passive party keeps going forward propagation, as shown in the step 1 of Sec~\ref{sec1}
It calculates the weighted sum  ($[\mathbf{z_i}]_c = {\mathbf{W_2}}\bigotimes[\mathbf{a_i}]_c$) and applies softmax. 
%
However, the non-linearity $e^{[\mathbf{z_i}]_c}$ cannot be computed in the ciphertext.
The solution to this problem is transmitting the weighted sum $[\mathbf{z_i}]_c$ back to the active party for decryption and get the plaintext $\mathbf{z_i}$ to calculate the softmax results~\cite{ijcai2018-547}. 
But directly sending the weighted sum $[\mathbf{z_i}]_c$ without any protection will leak the prediction to the active party. The active party can use the activation prediction pairs $(\mathbf{a_i,z_i})$ to learn the classifier part in the passive party. In this way, the active party can learn the weights of the model on the passive party and further approximate the label. To end this, instead of transmitting the weighted sum $[\mathbf{z_i}]_c$ directly back to the active party, the passive party will inject random noise to it $[\mathbf{z_i}]_c \bigoplus \epsilon_s$. The noise $\epsilon_s$ hides the real weighted sum, which prevents the active party from accumulating activation prediction pairs to infer the weights of the neural network on the passive party. 
%
Finally, the active party decrypts the noisy weighted sums$[\mathbf{z_i} + \epsilon_s]_c$ and sends the decrypted $\mathbf{z_i} + \epsilon_s$ to the passive party. And then the passive party removes the noise injected before and computes the prediction (softmax result). 

Another problem here is that the passive party get both $\mathbf{W_2}$ and $\mathbf{z_i}$ and can easily get $\mathbf{a_i}$ via linear regression. To end this, the passive party should use noisy weight ${\mathbf{\widetilde{W}_2}}$ to calculate the weighted sum  (${[\mathbf{\widetilde{z}_i}]_c} = {{\mathbf{\widetilde{W}_2}}}\bigotimes[\mathbf{a_i}]_c$) where ${\mathbf{\widetilde{W}_2}} = \mathbf{W_2} + \epsilon_{acc}$. 
Note that $\epsilon_{acc}$ is generated by the active party, we will discuss how to inject $\epsilon_{acc}$ to $\mathbf{W_2}$ in the Sec~\ref{sec4}. With same procedure, the passive party still need to add additional noise to ${[\mathbf{\widetilde{z}_i}]_c}$ and send ${[\mathbf{\widetilde{z}_i}+\epsilon_s]_c}$ back the active party to avoid computing non-linearity homomorphically. To perform the correct forward propagation the active party will cancel the noise in ${[\mathbf{\widetilde{z}_i}]_c}$ via $ {[\mathbf{{z_
i}}]_c} =  {[\mathbf{\widetilde{z}_i}]_c} -\epsilon_{acc}\mathbf{a_i}$ and send ${[\mathbf{{z_
i}}]_c}$ to the passive party for computing the true softmax output $\mathbf{p_i}$ with respect to $\mathbf{W_2}$. Since the passive party can only observe the noisy ${\mathbf{\widetilde{W}_2}}$ and ${[\mathbf{{z}}]_c}$ with respect to ${\mathbf{W_2}}$. 
The passive party cannot infer the activation $\mathbf{a_i}$. The whole forward propagation performs in a privacy-preserved manner.

\subsection{Secure Backward Propagation on the passive party}
\label{sec4}
During the backpropagation we need to estimate two gradient: the gradient $\frac{\partial \mathcal{L}_i}{\partial \mathbf{W_2}}$ and the gradient  $\frac{\partial \mathcal{L}_i}{\partial \mathbf{a_i}}$. Note that these two gradients are the linear transformation of either $\mathbf{a_i}$ or $\mathbf{W_2}$, both the active party and the passive party can derive what they want via regression. Not carefully dealing with gradient updating may cause a significant information leak.
In backpropagation, the passive party will compute the following gradients:
$\frac{\partial \mathcal{L}_i}{\partial \mathbf{p_i}_i} = \mathbf{t_i} - \mathbf{p_i}$, $[\frac{\partial \mathcal{L}_i}{\partial \mathbf{W_2}}]_c  = \frac{\partial \mathcal{L}_i}{\partial \mathbf{p_i}_i}\bigotimes [\mathbf{a_i}]_c$ and $\widetilde{\frac{\partial \mathcal{L}_i}{\partial \mathbf{a_i}}}  = \frac{\partial \mathcal{L}_i}{\partial \mathbf{p_i}} {\mathbf{\widetilde{W}_2}}$. 
Note that the gradient of weights $[\frac{\partial \mathcal{L}_i}{\partial \mathbf{W_2}}]_c$ is in the encrypted form. After weights updating $[\mathbf{W_2}]_c = \mathbf{W_2} - \eta [\frac{\partial \mathcal{L}_i}{\partial \mathbf{W_2}}]_c$, the weights $[\mathbf{W_2}]_c$ will result in the encrypted form. This causes the issue that in the forward propagation of the next iteration, there will be two encrypted quantities in calculating weighted sum $\mathbf{z_i} = [\mathbf{W_2}]_c [\mathbf{a_i}]_c $, which is incompatible with PHE. To avoid this situation, the passive party need to send the gradients of weight  $[\frac{\partial \mathcal{L}_i}{\partial \mathbf{W_2}}]_c$ to the active party and get the decrypted gradients  $\frac{\partial \mathcal{L}_i}{\partial \mathbf{W_2}}$ back. However, this solution is dangerous.
Since the passive party holds $\frac{\partial \mathcal{L}_i}{\partial \mathbf{p_i}_i}$ and the active party holds the $\mathbf{a_i}$, knowing the gradient $\frac{\partial \mathcal{L}_i}{\partial \mathbf{W_2}}$ make both parties leak information at the same time.  
Thus, both two parties need to add random noise to the gradients of weights before sending it to the other side. Specifically, they do 
\begin{itemize}
    \item The passive party add noise $\epsilon_s$ to $[\frac{\partial \mathcal{L}_i}{\partial \mathbf{W_2}}]_c$ and send $[\frac{\partial \mathcal{L}_i}{\partial \mathbf{W_2}} + \mathbf{\epsilon_s}]$ to the passive party.
    \item the active party decrypt $[ \frac{\partial \mathcal{L}_i}{\partial \mathbf{W_2}} + \epsilon_s]_c$
    \item The active party add random noise $\frac{{\epsilon_w}}{\eta}$:  
    $
        \widetilde{\frac{\partial \mathcal{L}_i}{\partial \mathbf{W_2}}}+ \epsilon_s \leftarrow \frac{\partial \mathcal{L}_i}{\partial \mathbf{W_2}} + \epsilon_s -\frac{{\epsilon_w}}{\eta}
    $
    \item The active party send  $\widetilde{\frac{\partial \mathcal{L}_i}{\partial \mathbf{W_2}}}+ \epsilon_s$ to the passive party
    \item The passive party remove noise $\epsilon_s$: 
    $
        \widetilde{\frac{\partial \mathcal{L}_i}{\partial \mathbf{W_2}}} \leftarrow \widetilde{\frac{\partial \mathcal{L}_i}{\partial \mathbf{W_2}}} + \epsilon_s
    $
\end{itemize}
Note that noise generated by the passive party can be removed immediately while the gradients of weight $\widetilde{\frac{\partial \mathcal{L}_i}{\partial \mathbf{W_2}}}$ still contain noise where $\widetilde{\frac{\partial \mathcal{L}_i}{\partial \mathbf{W_2}}}$ = ${\frac{\partial \mathcal{L}_i}{\partial \mathbf{W_2}}} - \frac{\epsilon_w}{\eta}$ .
With $\widetilde{\frac{\partial \mathcal{L}_i}{\partial \mathbf{W_2}}}$ the passive party blindly update the parameters as:
$$
    {{\mathbf{\widetilde{W}_2}}^{t+1}} = {\mathbf{W_2}^{t}} - (\eta {\frac{\partial \mathcal{L}_i}{\partial \mathbf{W_2}}} - \frac{\epsilon_w}{\eta})\Rightarrow 
    {{\mathbf{\widetilde{W}_2}}^{t+1}} = {\mathbf{W_2}^{t}} - \eta {\frac{\partial \mathcal{L}_i}{\partial \mathbf{W_2}}} + {\epsilon_w} \Rightarrow
    {{\mathbf{\widetilde{W}_2}}^{t+1}} = {\mathbf{W_2}^{t+1}} + {\epsilon_w}
$$
We can see that the noise $\epsilon_w$ will accumulate in weight $\mathbf{W_2}$ in each iteration. If we note the accumulated noise as $\epsilon_{acc} = \epsilon_w^1 + \epsilon_w^2 + \dots \epsilon_w^t$, the true weights that supposed to be used in the forward and backward propagation should be $ {{\mathbf{W_2}}^{t+1}} = {{\mathbf{\widetilde{W}_2}^{t+1}}} -\epsilon_{acc}$. To perform the right forward propagation, the active party needs to cancel the noise by subtracting $\mathbf{a_i}\epsilon_{acc}$ from the noisy weighted sums $\mathbf{\widetilde{z_i}}$ as described in Sec~\ref{sec3}. Similar to the forward propagation, in the back propagation, the extra noisy gradients is also added to $\widetilde{\frac{\partial \mathcal{L}_i}{\partial \mathbf{a_i}}}$ and need to be removed before backpropagating to the active party. To achieve this, the active party needs to send the encrypted $[\epsilon_{acc}]_c$ to the passive party. the passive party calculate the true gradient $[{\frac{\partial \mathcal{L}_i}{\partial \mathbf{a_i}}}]_c = \widetilde{\frac{\partial \mathcal{L}_i}{\partial \mathbf{a_i}}} - [\epsilon_{acc}]_c \bigotimes \frac{\partial \mathcal{L}_i}{\partial \mathbf{p_i}}$ and send the encrypted gradient $[{\frac{\partial \mathcal{L}_i}{\partial \mathbf{a_i}}}]_c$ to the active party.

\subsection{Secure Backward propagation on the active party}
The active party decrypt the gradient $[{\frac{\partial \mathcal{L}_i}{\partial \mathbf{a_i}}}]_c$ received from
the passive party, compute the gradient $\frac{\partial \mathcal{L}_i}{\partial \mathbf{W_1}} = \frac{\partial \mathcal{L}_i}{\mathbf{a_i}} \frac{\partial 
    \mathbf{a_i}}{\partial \mathbf{W_1}} = \frac{\partial \mathcal{L}_i}{\partial \mathbf{a_i}} \mathbf{x_i}$ 
    and update the parameters $\mathbf{W_1} = \mathbf{W_1} - \eta \frac{\partial \mathcal{L}_i}{\partial \mathbf{W_1}}$.
The whole privacy preserved backpropagation is detailed in Algorithm 1 and Algorithm 2.
\section{Experiment}
To evaluate the proposed method, specifically, we determine if our model (i)  achieves lossless performance in classification tasks and (ii) if it brings advantages over other solution based on learning on encrypted data. Please note we don't want to highlight the performance of neural networks but we can achieve a lossless performance. 
we implement the passive party and the active party both on two PCs with CPU Core i7-6850k and 64GB RAM connected by 1Gbps LAN. The neural network of the active party is built using pytorch~\cite{paszke2017automatic} and the neural net of the passive party is implemented by pure python integrated with $numpy$.
We compare the proposed method with two previous studies that have considered privacy-preserved training with homomorphic encryption for deep neural networks. GELU~\cite{ijcai2018-547} using PHE~\cite{paillier1999public} and CryptoNets~\cite{gilad2016cryptonets}. 

\subsection{Setting}
\label{sec:setting}
In order to make comparison with previous work: GELU and CryptoNets~\cite{gilad2016cryptonets}. We implement the following architecture of the proposed method:
\begin{itemize}
    \item \textbf {Multinomial Logistic Regression(MLR}: Data-Dense(10)-Softmax
    \item \textbf {Conv-1}: Data-Conv(5$\times$5, stride 2, 5 filter) -ReLu-MeanPooling-ReLu-Dense(84))-Dense(10)-Softmax
    \item \textbf {LeNet-5}:  Data-Conv(5$\times$5, stride 1, 6 filter)-MeanPooling-ReLu-Conv(5$\times$5, stride 1, 16 filter)-MeanPooling-ReLu-Dense(120)-Dense(84)-Dense(10)-Softmax
\end{itemize}
For the proposed method, the final layer, Dense(10)-Softmax, is on the passive party side, the rest of layers are on the active party side. 

\subsection{Training Accuracy}
In this section, we show the proposed method is a lossless solution. The neural network architecture we use is Conv-1. Due to arithmetic operations of homomorphic encryption only support multiplication and addition, CryptoNets uses the square function to avoid computing the non-linear activation. However, this makes the training unstable and damage to the accuracy~\cite{gilad2016cryptonets}. From the Table.~\ref{tab:tab1}, we can see that CryptoNets suffers an accuracy loss ranging from 2\% to 5\% compared with the proposed method. 
We also note that GELU is also a lossless solution. 
This is because GELU adopts the similar round trip strategy that sending activation back and forth between the passive party and the active party for handling non-linear operation as we do.
\begin{table}[htbp]
  \centering
  \caption{Test accuracy of different Architecture}
    \begin{tabular}{lccc}
    \toprule
    Architecture & \multicolumn{1}{l}{GELU} & \multicolumn{1}{l}{CryptoNets} &\multicolumn{1}{l}{Ours}\\
    \midrule
    Iris &0.982 &0.966&0.980\\
    \midrule
    Diabetes &0.758 &0.741 &0.760\\
    \midrule
    kr-vs-kp &0.967 &0.948 &0.965\\
    \midrule
    MNIST &0.969 &0.919 &0.970\\
    \bottomrule
    \end{tabular}
  \label{tab:tab1}%
\end{table}%

\subsection{Computation Speed}
As above mentioned, the dimension of inputs and the neural network architecture are key factors to other methods. In this section we will demonstrate that the proposed method is insensitive to these two factors.
%
\paragraph{Sensitivity of Input Dimension}
\begin{figure}
\centering
\includegraphics[width=0.5\textwidth]{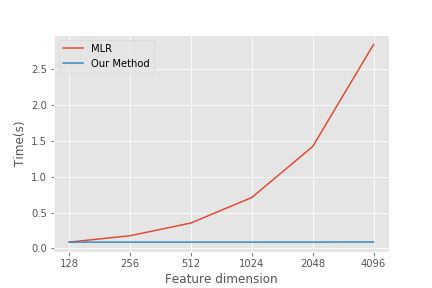}
\caption{The computation speed for one inference with different input feature dimension }
\label{fig:feature-dim} 
\end{figure}
One problem of learning on the encrypted data is the computation speed depend on the input dimension. It is because the dimension of the encrypted data directly increases the number of homomorphic encryption operations. We compare our model with multinomial logistic regression that learned on encrypted data.
We experiment with simulated data with various feature lengths. As shown in Fig.~\ref{fig:feature-dim}, our proposed method is insensitive to the dimensions of input data compared with multinomial logistic regression (i.e., shallow network with softmax activation). 
We do the experiment on the real word data using MNIST~\cite{lecun1998gradient} and CIFAR-10~\cite{krizhevsky2009learning}. We can observe from the table.~\ref{tab:tab2} the result of CIFAR-10 \cite{krizhevsky2009learning} (dimensions is $3\times32\times32$) is very similar to the result on MNIST \cite{lecun1998gradient} (dimensions is $1\times28\times28$). This benefits from we employ learning on encrypted features generated from the 'feature extraction' of the deep neural network, which can significantly reduce raw data dimensions. Compared with learning on encrypted data, feature extraction is very efficient. So that is why the proposed method consumes similar time in the different datasets.
\begin{table}[htbp]
\centering
    \caption{Time for different architectures on MNIST and CIFAR-10 dataset}
    \begin{tabular}{lr}
    \toprule
    Archiecture & \multicolumn{1}{l}{Time(s)} \\
    \midrule
    Multinomial Logistic Regression (MNIST)  & 0.5887 \\
    \midrule
    Multinomial Logistic Regression (CIFAR-10)  & 2.7027 \\
    \midrule
    LeNet-5 (MNIST) & 0.0583 \\
    \midrule
    LeNet-5 (CIFAR-10) & 0.0592 \\
    \bottomrule
    \end{tabular}
  \label{tab:tab2}%
\end{table}%

\paragraph{Sensitivity of Neural Network Architectures}
Note that the proposed method is designed to work on any architecture without much computation speed degrade. In comparison, the other baselines models like GELU and CryptoNet are sensitive to either the neural networks architecture or input data dimension. Thus, using complex neural net such as VGG~\cite{simonyan2014very}, ResNet~\cite{he2016deep} with high dimension input such as ImageNet~\cite{deng2009imagenet} is infeasible to the methods of learning on encrypted data, and it will take an extremely long time. Moreover, since the arithmetic operations in PHE is much faster than arithmetic operations in FHE. We mainly focus on comparing the proposed method with GELU. Thus, to make a fair comparison, we train the proposed method and GELU on different architecture described in Sec.~\ref{sec:setting}. We report the computation time (for one inference). 
\begin{table}[htbp]
  \centering
  \caption{Test accuracy of different Architectures}
    \begin{tabular}{lrr}
    \toprule
    Architecture & \multicolumn{1}{l}{Time(s)} & \multicolumn{1}{l}{Accuracy} \\
    \midrule
    GELU-Net(LeNet-5) & 7.855 & 0.989 \\
    \midrule
    Multinomial Logistic Regression & 0.588 & 0.93 \\
    \midrule
    Our Method(LeNet-5) & 0.0583 & 0.99 \\
    \bottomrule
    \end{tabular}
  \label{tab:tab3}%
\end{table}%
\begin{table}[htbp]
  \centering
  \caption{Time for the proposed method and GELU-Net on MNIST}
    \begin{tabular}{lr}
    \toprule
    Archiecture & \multicolumn{1}{l}{Time(s)} \\
    \midrule
    GELU-Net(Conv-1) & 3.8437 \\
    \midrule
    Our Method (Conv-1)  & 0.0582 \\
    \midrule
    Our Method(LeNet-5)  & 0.0583 \\
    \bottomrule
    \end{tabular}%
  \label{tab:tab4}%
\end{table}%
Table.\ref{tab:tab3} shows the computation time (for one inference) and the accuracy on MNIST \cite{lecun1998gradient}. We observe that the proposed method achieves more than 100x speed-up over GELU-Net with no accuracy loss. Even compared with simpler architecture multinomial logistic regression (MLR), our method is much faster. Therefore our method works much faster than GELU and performs much better than the linear model with encrypted data due to the deep neural network can learning more semantic represendation for targets.
Table.~\ref{tab:tab4} shows the computation speed of the proposed method in both deep (LeNet-5) and shallow neural net (Conv-1). Result reveals that the time for one inference is almost invariant to the architectures as long as the output dimension is the same (both Conv-1 and LeNet-5 are 84). That means the mainly computational bottleneck is on the 'classifier' which learns the part model on encrypted data. It also shows that our deep model is highly potential to apply to more complicated data which need the deeper neural network.

\section{Conclusion}
In this paper, we have proposed a scheme, called asymmetrically collaborative machine learning where one party has data, but the other party has labels only.
A deep neural network with a partly unencrypted and partly encrypted strategy is proposed to avoid learning on encrypted data directly for this scheme.
Beyond that, we offer a series of solutions to preserve privacy from both parties involved.
The design has ensured the efficiency and effectiveness of the proposed method. 
We have carried out extensive experiments that demonstrate more than $100\times$ times speedup compared with the state-of-the-art solutions.

\bibliographystyle{plain}
\bibliography{ref}

\end{document}